\documentclass{article}
\usepackage{enumitem}
\usepackage{algorithm}
\usepackage{algpseudocode}
\usepackage{graphicx}   
\usepackage{float}      
\usepackage{placeins} 
\usepackage{listings}
\usepackage{xcolor} 
\usepackage{amssymb} 
\usepackage{verbatimbox}
\usepackage{url}

\usepackage[utf8]{inputenc}  
\usepackage{graphicx}
\usepackage[english]{babel}

\usepackage[letterpaper,top=2cm,bottom=2cm,left=3cm,right=3cm,marginparwidth=1.75cm]{geometry}

\usepackage{amsmath}
\usepackage{graphicx}
\usepackage[colorlinks=true, allcolors=blue]{hyperref}
\usepackage[utf8]{inputenc}
\usepackage{hyperref}
\usepackage{url}
\usepackage{centernot}
\usepackage{geometry}

\title{RAGulating Compliance: A Multi-Agent Knowledge Graph for Regulatory QA}
\author{
  Bhavik Agarwal, Hemant Sunil Jomraj, Simone Kaplunov, Jack Krolick, Viktoria Rojkova \\
  MasterControl AI Research \\
  \texttt{\{bagarwal,hjomraj,skaplunov, jkrolick,vrojkova\}@mastercontrol.com}
}
\date{}

\begin{document}
\maketitle

\maketitle

\begin{abstract}
Regulatory compliance question answering (QA) requires precise, verifiable information, and domain-specific expertise, posing challenges for Large Language Models (LLMs). In this work, we present a novel multi-agent framework that integrates a Knowledge Graph (KG) of Regulatory triplets with Retrieval-Augmented Generation (RAG) to address these demands. First, agents build and maintain ontology-free KG by extracting subject–predicate–object (SPO) triplets from regulatory documents and systematically cleaning, normalizing, deduplicating, and updating them. Second, these triplets are embedded and stored along with their corresponding textual sections and metadata in a single enriched vector database, allowing for both graph-based reasoning and efficient information retrieval. Third, an orchestrated agent pipeline leverages triplet-level retrieval for question answering, ensuring high semantic alignment between user queries and the factual 'who-did-what-to-whom' core captured by the graph. Our hybrid system outperforms conventional methods in complex regulatory queries, ensuring factual correctness with embedded triplets, enabling traceability through a unified vector database, and enhancing understanding through subgraph visualization, providing a robust foundation for compliance-driven and broader audit-focused applications. 
\end{abstract}

\section{Introduction}

The growing regulatory complexities in healthcare, pharmaceuticals, and medical devices shape market access and patient care \cite{YuHan2024}. The extensive guidance and rules of the FDA \cite{FDA2025} require strict compliance with approvals, post-market surveillance, and quality systems \cite{Cordes2022}. Meanwhile, LLMs such as GPT-o1 \cite{Zhong2024}, Qwen-2.5 \cite{Yang2024} and Pi-4 \cite{Abdin2024} excel in text tasks but face unique challenges in precision, verifiability, and domain specialization in high-stakes regulatory contexts \cite{Wang2024}. Hallucination risks and limited contextual understanding underscore the need for robust guardrails, particularly in safety-critical applications \cite{Hakim2024}, \cite{Ling2024}. How can we ensure the domain specificity and reliability required for compliance?

Our work proposes a three-fold innovation for regulated compliance: first, we construct and refine triplet graphs from regulatory documents, building on knowledge graph research \cite{Nickel2015}; second, we integrate these graphs with RAG techniques, inspired by open-domain QA \cite{Lewis2021} and healthcare question-answering \cite{Yang2025}, to reduce hallucinations \cite{Ji2024}; and third, a multi-agent architecture oversees graph construction, RAG database enrichment, and the final question-answering process, ultimately grounding responses in factual relationships to enhance precision, reliability, and verifiability—key for demonstrating compliance to regulators and stakeholders.

\section{Relevant Work}
A line of research tackles hallucination and domain-specific gaps by integrating language models with knowledge graphs (KGs), which encode domain knowledge for semantic linking, inference, and consistency checks \cite{Hogan2021,Nickel2015,Chen2020}. In regulatory settings, KGs capture complex relationships among rules and guidelines \cite{Chattoraj2024}, and when combined with retrieval-augmented generation (RAG) \cite{Lewis2021}, reduce factual errors by putting outputs in authoritative data \cite{Li2024}. Although RAG has excelled in open-domain QA \cite{Lewis2021,Karpukhin20}, its application in regulatory compliance, particularly synthesizing structured (KG) and unstructured text, remains underexplored. Multi-agent systems \cite{Shoham08,Wooldridge09} offer autonomous agents for data ingestion, KG construction, verification, and inference, enabling modularity and scalability \cite{Weiss00,Zygmunt13}. This approach is well suited to dynamic regulatory environments that require constant updates.

\subsection{Knowledge Graphs in Regulatory Compliance} Knowledge graphs excel at representing complex regulatory information, facilitating semantic relationships \cite{Hogan2021}. Notable examples include enterprise KGs for market regulations \cite{Ershov23} and frameworks for medical device policies \cite{Chattoraj2024}, while \cite{Xiang25} underscores KG reasoning techniques that bridge structured and unstructured data.

\subsection{Retrieval-Augmented Generation in Regulatory Compliance} RAG \cite{Lewis2021} integrates retrieval mechanisms with generative language models, improving the factual accuracy \cite{Hillebrand24}. In the pharmaceutical domain, a chatbot that uses RAG successfully navigated complex guidelines by retrieving and storing responses in relevant documents \cite{Kim24}.

\subsection{Multi-Agent Systems and Their Application} Multi-agent systems enable specialized agents to coordinate complex tasks \cite{Shoham08,Wooldridge09}, facilitating robust data integration and knowledge engineering \cite{Zygmunt13}—a key advantage in rapidly evolving regulatory contexts.

\section{Ontology Free Knowledge Graph}
\begin{figure}[h]
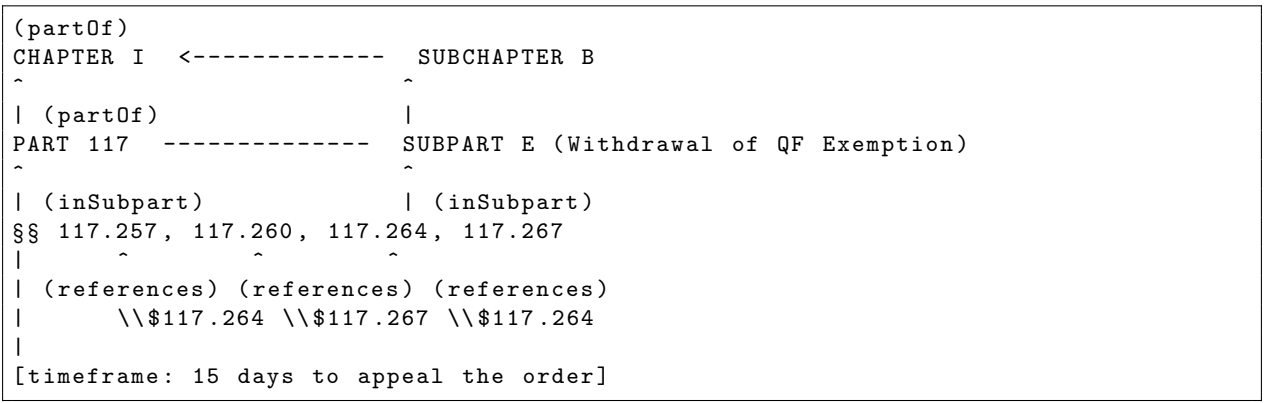

\centering
\begin{lstlisting}[basicstyle=\ttfamily\small, frame=single, numbers=none]
(partOf)
CHAPTER I  <-------------  SUBCHAPTER B
^                         ^
| (partOf)                |
PART 117  --------------  SUBPART E (Withdrawal of QF Exemption)
^                         ^
| (inSubpart)             | (inSubpart)
§§ 117.257, 117.260, 117.264, 117.267
|      ^        ^        ^
| (references) (references) (references)
|      \\$117.264 \\$117.267 \\$117.264
|
[timeframe: 15 days to appeal the order]
\end{lstlisting}
\caption{Independent sections converge on a single requirement, only discernible through triplet-driven interconnections.}
\label{fig:ontology_free_schematic}
\end{figure}

Knowledge graphs often rely on predefined ontologies (e.g. DBpedia \cite{Lehmann15}, YAGO \cite{suchanek2007yago}), yet an alternative 'schema-light' approach defers rigid schemas in favor of flexible bottom-up extraction \cite{etzioni2011open, fader2014open}. This method quickly adapts to new data domains \cite{carlson2010architectureNELL}, reduces initial overhead \cite{etzioni2011open}, and allows partial schemas to emerge naturally \cite{hogan2021knowledgegraphs}, making it especially valuable in regulatory settings where rules evolve rapidly, data formats vary \cite{probst2006scalable}, and open-ended queries can reveal hidden legal connections \cite{fader2014open}.
In order to demonstrate how the \emph{schema-light} strategy operates in practice, we extracted triplets from the data from the Electronic Code of Federal Regulations, focusing on specific sections that share references and time constraints. The resulting relationships form a small subgraph that illustrates both shallow hierarchical structures (\emph{e.g.}, parts, and subparts) and interlinked regulatory requirements. As seen in Figure~\ref{fig:ontology_free_schematic}, these extracted triplets reveal how different sections converge on the same 15-day appeal timeframe, underscoring the flexibility of an ontology-free approach in capturing cross-references and shared procedural deadlines.

\section{Triplet-Based Embeddings for Regulatory QA with Textual Evidence}

In this section, we introduce a formulation for leveraging embedded triplets to enable precise, fact-centric retrieval in a regulatory question-answering system. Unlike purely text-based approaches, our method not only encodes concise Subject-Predicate-Object relationships, but also links each triplet to the original text sections from which it was extracted. At query time, the system retrieves both the relevant triplets and corresponding text evidence, feeding them into an LLM for the final generation of the answer.
\subsection{Corpus, Sections, and Triplet Extraction}

Let $\mathcal{C}$ be a corpus of regulatory documents. We partition $\mathcal{C}$ into atomic text sections ---- for instance, paragraphs, clauses, or semantically coherent fragments ----using a function
\[
\Omega: \mathcal{C} \;\to\; \mathcal{X},
\]
where $\mathcal{X}$ is the set of all text fragments
\(\{\,x_{1},\,x_{2},\,\dots,\,x_{m}\}\).

We then apply an information extraction pipeline $\Phi$ to each section \(x_j\). The pipeline identifies the subject-predicate-object relationships of that section, thus producing triplets:
\[
\Phi\bigl(\Omega(\mathcal{C})\bigr)
\;=\;
\bigl\{\,
t_i
\;\mid\;
t_i = (s_i,\; p_i,\; o_i)
\bigr\}.
\]

We define a linking function $\Lambda$ such that each triplet \(t_i\) is associated with one or more text sections \(x_j\). Formally,
\[
\Lambda:
\mathcal{T}
\;\to\;
2^{\mathcal{X}},
\]
where $\mathcal{T}$ is the set of all extracted triplets, and $\Lambda\bigl(t_i\bigr)$ yields the subset of sections from which $t_i$ was extracted. Hence, each triplet $t_i$ has \emph{provenance}---a reference to its original textual source(s).
\subsection{Embedding Triplets}

For each triplet \( t_i = (s_i,\, p_i,\, o_i) \), we create a short textual representation \(f(t_i)\). A typical choice is a concatenation of S-P-O, for example:
\[
f(t_i) = \mathrm{concat}(s_i,\; p_i,\; o_i).
\]
We then define an embedding function
\[
E: \mathcal{X} \cup \mathcal{T} \;\to\; \mathbb{R}^{d},
\]
where \(d\) is the dimensionality of the embedding space. Specifically, for any triplet \(t_i\),
\[
\mathbf{e}_{t_i} = E\bigl(f(t_i)\bigr) \;\in\; \mathbb{R}^d
\]

We also embed queries and (optionally) text sections themselves via the same or a compatible model. The resulting vectors are stored in a vector index \(\mathcal{V}\) such that:
\[
\mathcal{V}
=
\bigl\{
\, (\mathbf{e}_{t_i},\, t_i,\, \Lambda(t_i))
\;\bigm|\; 1 \,\le\, i \,\le\, N
\bigr\},
\]
where \(\mathbf{e}_{t_i}\) is the triplet embedding and \(\Lambda(t_i)\) is the set of associated text sections.

\subsection{Embedding Function}
To enhance query processing and retrieval, we developed an embedding model based on Transformer's methodology, specifically leveraging transformer-based architectures such as BERT. This embedding model was trained on textual data extracted from the eCFR, capturing semantic nuances specific to the regulatory language. The embedding process involves encoding cleaned textual chunks into high-dimensional vector representations, which enable efficient semantic search and retrieval in downstream tasks, significantly improving the precision and relevance of responses to regulatory queries.

\subsection{Query Embedding and Retrieval}

Given a user query \(Q \in \mathcal{Q}\) ---- for example, ``Which agency is responsible for Regulation 2025-X?'' ---- we embed \(Q\) as
\[
\mathbf{e}_Q = E(Q).
\]
We perform a \(k\)-nearest neighbor search in \(\mathcal{V}\) using a similarity measure \(\mathrm{sim}(\cdot,\cdot)\), typically cosine similarity. We obtain:
\[
\mathcal{T}_Q = \mathrm{TopK} \Bigl(
\mathrm{sim}\bigl(\mathbf{e}_Q, \; \mathbf{e}_{t_i}\bigr) \Bigr)
\]

which yields the top-\(k\) triplets most relevant to the query. For each retrieved triplet \(t_i \in \mathcal{T}_Q\), we can also retrieve its associated text sections through \(\Lambda(t_i)\). Formally:
\[
\mathcal{X}_Q = \bigcup_{t_i \,\in\, \mathcal{T}_Q} \Lambda\bigl(t_i\bigr)
\]

so \(\mathcal{X}_Q\) is the set of sources' text sections that support the discovered triplets.

\subsection{LLM-Based QA with Triplets and Text}

To finalize the answer, we define a function
\[
\Gamma: \mathcal{Q} \;\times\; 2^{\mathcal{T}} \;\times\; 2^{\mathcal{X}}
\;\to\;
\mathcal{A},
\]
where \(\mathcal{A}\) is the set of possible answers. Essentially, \(\Gamma\) is an LLM that accepts: User query \(Q\), Retrieved triplets \(\mathcal{T}_Q\), Relevant Text Sections \(\mathcal{X}_Q\).
The LLM then produces an answer \(A \in \mathcal{A}\). In symbolic form:
\[
A = \Gamma\bigl(Q,\; \mathcal{T}_Q,\; \mathcal{X}_Q\bigr).
\]
In practice, the LLM input might be a prompt that includes the user question plus concatenated or selectively summarized triplets and text sections. By examining both structured (triplet) facts and verbatim textual evidence, the LLM generates a more accurate and explainable response.

\subsection{Theoretical Considerations}
\paragraph{}\textbf{Completeness and Consistency:} \(\mathcal{T}\) is complete if every relevant statement in \(\mathcal{C}\) is represented by at least one SPO triplet and consistent if \(\Phi\) does not introduce contradictory or spurious triplets.
\paragraph{}\textbf{Retrieval Sufficiency:}
With \(\mathrm{sim}\bigl(\mathbf{e}_Q, \mathbf{e}_{t_i}\bigr)\) as a semantic relatedness measure and an embedding function \(E\) that preserves factual relationships, the top-\(k\) triplets in \(\mathcal{T}_Q\) should suffice to answer \(Q\).
\paragraph{}\textbf{Text Sections as Evidence:}
Because each \(t_i\) links back to its source text, users or downstream models can verify and clarify relationships by referring to the original regulatory language, thus mitigating ambiguities not fully captured by the triplet alone.

\section{Multi Agents System}
We use a multiagent system to orchestrate ingestion, extraction, cleaning, and query-answering in a modular, scalable manner. Each agent specializes in a core function, such as document ingestion, triplet extraction, or final answer generation, so they can run independently and be refined without disrupting the rest. 
\begin{figure}[H]
  \centering
  \includegraphics[width=16cm, height=1.5cm]{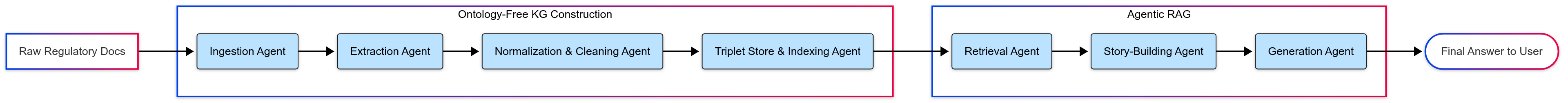}
  \caption{Multi Agents High Level Architecture}
\end{figure}

\subsection{Agents for ontology free knowledge graph constructions}

The document ingestion agent segments raw regulatory text, captures metadata, and outputs structured fragments. The extraction agent uses an LLM to detect subject-predicate-object triplets (e.g., 'FDA requires submission within 15 days'). Normalization and Cleaning Agent merges duplicates, standardizes entities, and resolves synonyms to produce clean triplets. Triplet Store and Indexing Agent embeds and stores triplets in a vector database for easy retrieval.

\subsection{Agentic Retrieval-Augmented Generation System}

Our second agentic system utilizes the custom embedding model to retrieve semantically similar triplets from the knowledge graph. Initially, the retrieval agent identifies relevant triplets based on semantic proximity to user queries. Subsequently, the story-building agent compiles and synthesizes the textual chunks associated with these triplets into a coherent narrative. Finally, the generation agent processes this cohesive story to formulate precise and contextually relevant responses. This approach ensures that responses to regulatory inquiries are accurate, traceable, and grounded in verified regulatory content.

\section{Retrieved Subgraph Visualization }
Additionally, we supplement the responses with an interactive visual of the relevant subgraphs of the retrieved triplets. This visual aid significantly improves user comprehension and provides greater contextual clarity, facilitating informed decision making in regulatory compliance tasks.
\begin{figure}[H]
  \centering
  \includegraphics[width=16cm, height=12cm]{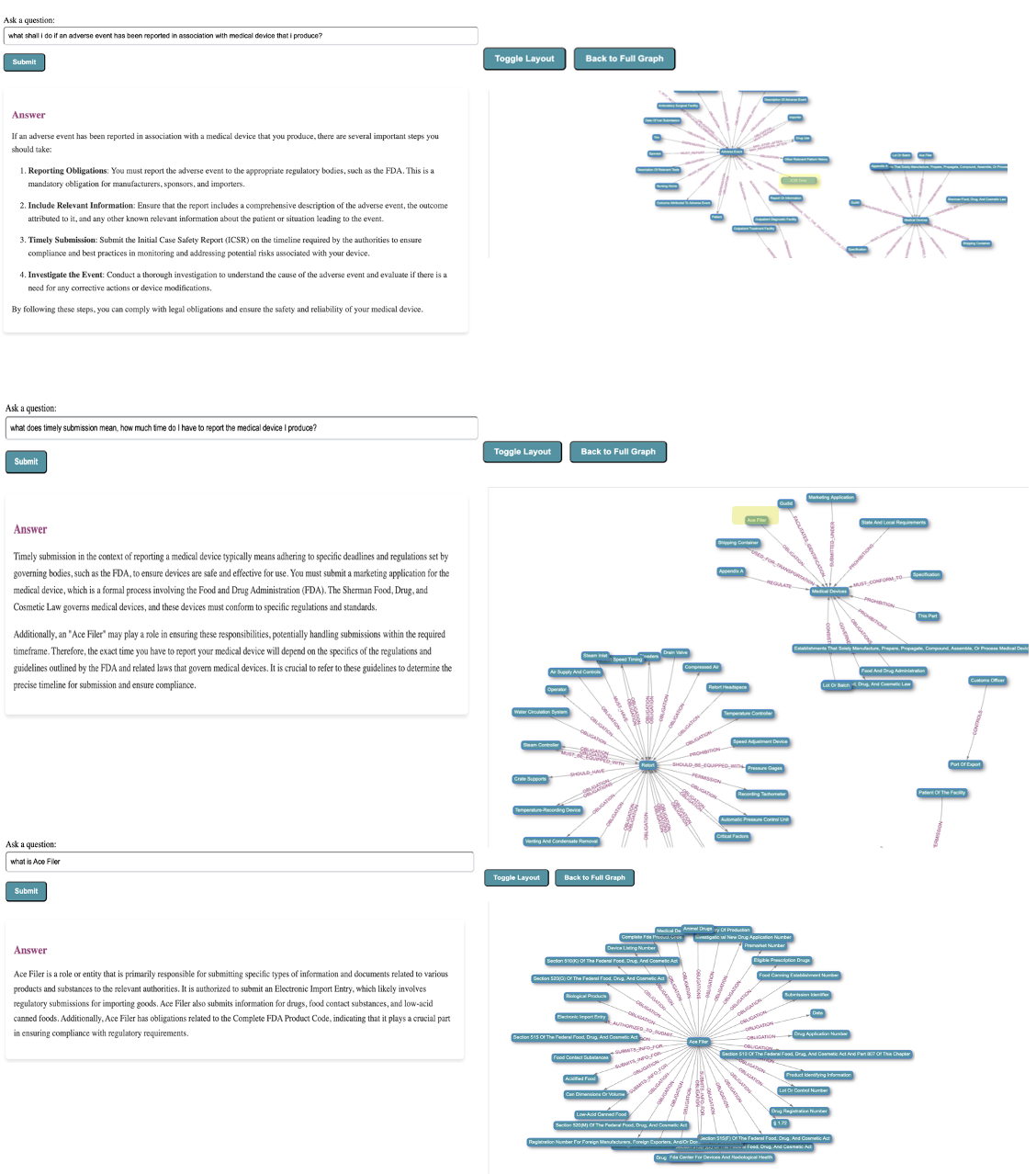}
  \caption{Navigational Facility of triplets}
\end{figure}

\section{Evaluation}
In this section, we outline our methodology for evaluating the system’s ability to (1) retrieve the correct sections of a regulatory corpus, (2) generate factually accurate answers and (3) demonstrate flexibility of navigation through the interconnection of triplets in related sections. We detail our sampling procedure, the construction of queries, the measurement of section-level overlap, the assessment of factual correctness, and the analysis of triplet-based navigation.
\begin{figure}[H]
  \centering
  \includegraphics[width=16cm, height=2cm]{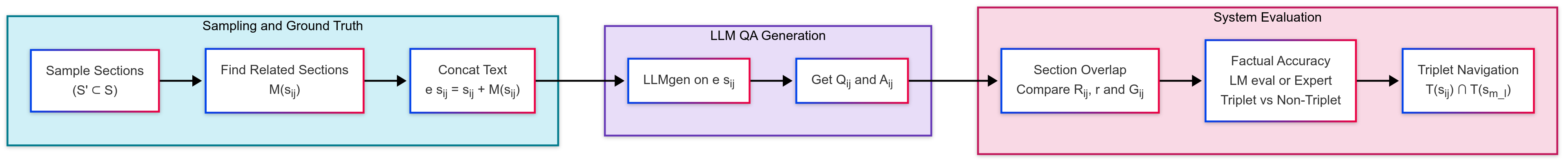}
  \caption{Evaluation Methodology}
\end{figure}

\subsection{Sampling and Ground Truth Construction}

\paragraph{Random Sampling of Sections.}
Let \(\mathcal{S} = \{\,s_1,\, s_2,\, \dots,\, s_N\}\) be the full set of sections of the regulatory corpus. We draw a random subset
\[
\mathcal{S}' \;=\; \bigl\{\, s_{i_1},\, s_{i_2},\, \dots,\, s_{i_k}\bigr\}
\subset \mathcal{S},
\]
where each \(s_{i_j}\) is considered a \emph{target section} for evaluation, and \(k \ll N\).

\paragraph{Identifying All Ground Truth Mentions.}
For each sampled section \(s_{i_j}\), we locate all other sections in the corpus that reference or expand upon the same regulatory ideas or entities. Formally, let
\[
M(s_{i_j}) \;=\; \bigl\{\, s_{m_1},\, s_{m_2},\, \dots \bigr\}
\]
denote the set of sections that contain overlaps or references relevant to \(s_{i_j}\). We then create a \emph{re-told story} by concatenating \(s_{i_j}\) with all sections in \(M(s_{i_j})\):
\[
\widetilde{s}_{i_j}
\;=\;
\bigl(\,
s_{i_j}
\;\Vert\;
s_{m_1}
\;\Vert\;
s_{m_2}
\;\Vert\;\dots
\bigr).
\]
This concatenated text \(\widetilde{s}_{i_j}\) is treated as the ground truth context for the focal section \(s_{i_j}\).

\subsection{LLM-Generated Questions and Answers}

We employ a Large Language Model, denoted \(\mathrm{LLM}_{\text{gen}}\), to produce a set of questions and corresponding reference answers based on each concatenated text \(\widetilde{s}_{i_j}\). Formally,
\[
(\mathcal{Q}_{i_j},\, \mathcal{A}_{i_j})
\;=\;
\mathrm{LLM}_{\text{gen}}\bigl(\,\widetilde{s}_{i_j}\bigr),
\]
where \(\mathcal{Q}_{i_j} = \{q_1, q_2, \dots, q_m\}\) and \(\mathcal{A}_{i_j} = \{a_1, a_2, \dots, a_m\}\). Each pair \((q_r, a_r)\) is presumed to be responsible via the original information in \(\widetilde{s}_{i_j}\).

\subsection{System Inference and Evaluations}

\subsubsection{Section-Level Overlap}

To answer each question \(q_r \in \mathcal{Q}_{i_j}\), our system retrieves a set of sections \(\mathcal{R}_{i_j,r}\) deemed relevant (based on embedding retrieval, triplet matching, or both). We measure the \textbf{level of overlap} between the recovered sections \(\mathcal{R}_{i_j,r}\) and the ground truth target section \(s_{i_j}\) (along with its reference set \(M(s_{i_j})\)).

\paragraph{Definition: Overlap score.}
Let \(\mathcal{G}_{i_j} = \{s_{i_j}\} \cup M(s_{i_j})\) be the set of ground truth sections. Suppose that the system returns \(\mathcal{R}_{i_j,r} = \{r_1, r_2, \dots, r_\ell\}\). We define the overlap score \(\mathcal{O}\) for question \(q_r\) as
\[
\mathcal{O}\bigl(\mathcal{R}_{i_j,r},\, \mathcal{G}_{i_j}\bigr)
\;=\;
\frac{\bigl|\mathcal{R}_{i_j,r} \,\cap\, \mathcal{G}_{i_j}\bigr|}
{\bigl|\mathcal{R}_{i_j,r}\bigr|}.
\]
Thus, 
\begin{itemize}
\item if \(\mathcal{R}_{i_j,r}\cap\mathcal{G}_{i_j} = \emptyset\), then \(\mathcal{O} = 0\);
\item if \(\mathcal{R}_{i_j,r}\) returns exactly one section, \(r_1\), and \(r_1 = s_{i_j}\), then \(\mathcal{O} = 1\);
\item if, for instance, the system returns three sections, only one of which matches any in \(\mathcal{G}_{i_j}\), then \(\mathcal{O} = 1/3\).
\end{itemize}

We can further refine this measure by applying a similarity threshold \(\theta\) for the equivalence between the retrieved sections and the ground truth sections (e.g., if the sections partially overlap or are highly similar). In that case,
\[
\bigl|\mathcal{R}_{i_j,r} \,\cap\, \mathcal{G}_{i_j}\bigr|
\;=\;
\sum_{r \in \mathcal{R}_{i_j,r}}
\mathbf{1}\Bigl[\mathrm{sim}\bigl(r,\, s_{g}\bigr) \,\ge\, \theta
\;\text{for some}\; s_g \in \mathcal{G}_{i_j}\Bigr].
\]

\subsubsection{Factual Correctness of Answers}

Once the system retrieves relevant sections and processes them through the QA pipeline (with or without associated triplets), it produces an answer \(a_r^\star\). We compare \(a_r^\star\) with the reference answer \(a_r\) from \(\mathrm{LLM}_{\text{gen}}\).

\paragraph{LLM-Based Fact Checking.}
We use a secondary evaluation model \(\mathrm{LLM}_{\text{eval}}\) or a domain expert to assess whether \(a_r^\star\) is \emph{factually correct} with respect to the original text \(\widetilde{s}_{i_j}\). We denote:
\[
F(a_r^\star,\; a_r) \;=\;
\begin{cases}
1, & \text{if $a_r^\star$ is factually correct and consistent with $a_r$}, \\
0, & \text{otherwise}.
\end{cases}
\]
We measure correctness with two conditions:
\begin{enumerate}
\item \textbf{With Triplets}: The system’s answer is grounded in the set of triplets that directly link to the retrieved sections.
\item \textbf{Without Triplets}: The system response is derived purely from the retrieval of raw text, without referencing the triplet data structure.
\end{enumerate}
By comparing the correctness scores for these two conditions, we quantify the \emph{impact of structured triplets} in factual precision.

\subsubsection{Navigational Facility of Triplets}

We also investigate how \emph{triplet interconnections} facilitates follow-up questions. In many regulatory contexts, a concept from one section leads to further questions about a related section. To do this, we define the following.

\paragraph{Triplet Overlap Across Sections.}
Let \(\mathcal{T}\) be the global set of extracted triplets. For sections \(s_{i_j}\) and \(s_{m_\ell}\in M(s_{i_j})\), we look at triplets that are shared or linked between these sections:
\[
\mathcal{T}(s_{i_j}) \;=\; \{\, t\in \mathcal{T}\mid t\ \text{is extracted from section}\ s_{i_j}\},
\]
\[
\mathcal{T}(s_{m_\ell}) \;=\; \{\, t\in \mathcal{T}\mid t\ \text{is extracted from section}\ s_{m_\ell}\}.
\]
We then analyze:
\[
\mathcal{T}(s_{i_j}) \,\cap\, \mathcal{T}(s_{m_\ell}),
\]
which denotes shared triplets that link the heads / tail entities in sections. A single triplet may appear in multiple sections if those sections refer to the same entity relationships; or it may connect a head entity in \(s_{i_j}\) to a tail entity in \(s_{m_\ell}\).

\paragraph{Navigational Metric.}
We define a metric \(\mathrm{Nav}(\mathcal{S}')\) to capture \emph{average fraction of shared or sequentially linked triplets} among sections that mention the same ground-truth concepts. Let
\[
\mathrm{Nav}(\mathcal{S}')
\;=\;
\frac{1}{k}\,\sum_{j=1}^k
\frac{\sum_{\,s_{m_\ell}\in M(s_{i_j})}
\bigl|
\mathcal{T}(s_{i_j}) \,\cap\, \mathcal{T}(s_{m_\ell})
\bigr|}
{\sum_{\,s_{m_\ell}\in M(s_{i_j})}
\bigl|
\mathcal{T}(s_{i_j}) \,\cup\, \mathcal{T}(s_{m_\ell})
\bigr|}.
\]
A higher value indicates stronger overlap (and thus \emph{navigational facility}), suggesting that triplets help the system move seamlessly between related sections.

By integrating section-level overlap analysis, factual correctness checks, and a triplet interconnection navigation metric, this evaluation framework measures retrieval accuracy, answer precision, and knowledge connectivity - ensuring robust compliance support, domain-specific Q\&A, and effective scalability in real-world regulatory settings.

\begin{table}[htbp]
    \centering
    \caption{Evaluation Results for Section Overlap, Answer Accuracy, and Navigation Metrics}
    \label{tab:results}
    \begin{tabular}{|l|c|c|}
        \hline
        \textbf{Metric} & \textbf{Without Triplets} & \textbf{With Triplets} \\ 
        \hline
        \multicolumn{3}{|l|}{\textbf{1. Section Overlap (Similarity Threshold)}} \\ 
        \hline
        0.50 & 0.0812 & 0.0745 \\ 
        0.60 & 0.2700 & 0.2143 \\ 
        \textbf{0.75 (stricter)} & 0.1684 & \textbf{0.2888 (highest accuracy)} \\ 
        \hline
        \multicolumn{3}{|l|}{\textbf{2. Answer Accuracy (Scale: 1-5)}} \\ 
        \hline
        Average Accuracy & 4.71 & \textbf{4.73} \\ 
        \hline
        \multicolumn{3}{|l|}{\textbf{3. Navigation Metrics}} \\ 
        \hline
        Average Degree & 1.2939 (less interconnected) & \textbf{1.6080 (more interconnected)} \\ 
        Unconnected Sections Linked & 5014 unconnected section & \textbf{5011 connected sections} \\ 
        Avg. Shortest Path & 2.0167 (slower information flow) & \textbf{1.3300 (faster information flow)} \\ 
        \hline
    \end{tabular}
        \vspace{1ex}
    \begin{minipage}{0.95\textwidth}
        \small
        The table compares system performance with and without triplets across three evaluation criteria: retrieval accuracy (section overlap at varying similarity thresholds), factual correctness of generated answers, and efficiency of navigation through related regulatory sections. Triplets yield highest accuracy at higher threshold. Triplets network significantly enhances connectivity and navigation.
    \end{minipage}
    
\end{table}

\section{Discussion}

Throughout this work, we presented a multi-agent system that uses triplet-based knowledge graph construction and retrieval-augmented generation (RAG) to enable transparent, verifiable question-answering on a regulatory corpus. By delegating ingestion, triplet extraction, KG maintenance, and query orchestration to specialized agents, unstructured text becomes a structured data layer for precise retrieval. The synergy of KG and RAG provides high-confidence, explainable facts alongside fluent responses to the large language model, as Section 7 demonstrates through accurate section retrieval, factual correctness and navigational queries (Figure~3). Grounding answers with triplets reduces LLM hallucinations, and provenance links enable robust auditing.

\subsection{Challenges}
An ontology-free approach facilitates rapid ingestion and incremental refinement but can lead to vocabulary fragmentation; canonicalization and entity resolution \cite{galarraga2014canonicalizing, shen2014entity} help unify concepts, and advanced reasoning tasks may still benefit from partial or emergent schemas \cite{riedel2013relation}. Extraction quality directly affects the integrity of the KG, as domain-specific jargon or ambiguous references can produce missing or erroneous triples, and deeper inferences or temporal constraints may require additional rule-based or symbolic logic. Large-scale RAG pipelines also require careful optimization for embedding, indexing, and retrieval.

\subsection{Future Directions}
Looking ahead, we see multiple avenues for enhancing and extending the system: although our current pipeline supports factual lookups, more complex regulatory questions demand deeper logical reasoning or chaining of evidence, and integration with advanced reasoning LLMs can address multistep analysis and domain-specific inference needs. 
By including user feedback or expert annotations, we could iteratively refine triplet quality and reduce extraction errors. Active learning or weakly supervised methods may help identify ambiguous relationships, prompting relabeling or model retraining. Over time, such feedback loops would yield higher-precision knowledge graphs.
Regulatory corpora often change rapidly (e.g., new guidelines, amendments). We aim to develop \emph{incremental update mechanisms} that re-ingest altered documents and regenerate only those triples affected by the changes, minimizing downtime and ensuring continuous compliance coverage.
Although we focus on \emph{health life science regulatory compliance}, the underlying architecture of multi--agent ingestion, knowledge graph construction, and RAG QA---can be generalized to other domains with high stakes factual queries (e.g., clinical trials, financial regulations, or patent law). Tailoring the extraction logic and graph schema of each agent to domain-specific requirements would enable a larger impact.

\bibliographystyle{alpha}
\bibliography{sample}

\end{document}